\documentclass[10pt,twocolumn,letterpaper]{article}

\usepackage{iccv}      
\usepackage[dvipsnames]{xcolor}
\usepackage[accsupp]{axessibility}
\usepackage{amsmath}
\usepackage{amssymb}
\usepackage{booktabs}
\usepackage{graphicx} 
\usepackage{subfloat} 
\usepackage{multirow}
\usepackage[whole]{bxcjkjatype}

\newcommand{\Tref}[1]{Table~\ref{#1}}
\newcommand{\eref}[1]{Eq.~\eqref{#1}}

\newcommand{\fref}[1]{Fig.~\ref{#1}}
\newcommand{\Fref}[1]{Figure~\ref{#1}}
\newcommand{\sref}[1]{Sec.~\ref{#1}}

\usepackage{xcolor}
\newcounter{todos}
\AtEndDocument{\ifnum\value{todos}>0 \PackageWarning{TODOS}{There are \arabic{todos} todos left in this paper! Fix them before submitting the paper!} \fi}

\newcommand{\V}[1]{\ensuremath{\mathbf{#1}}}

\usepackage{bm}

\newcommand{\real}{\mathbb{R}}
\newcommand{\modelname}{NeuraLeaf\xspace}
\newcommand{\datasetname}{DeformLeaf\xspace}

\definecolor{iccvblue}{rgb}{0.21,0.49,0.74}
\usepackage[pagebackref,breaklinks,colorlinks,allcolors=iccvblue]{hyperref}

\def\paperID{7665} 
\def\confName{ICCV}
\def\confYear{2025}

\begin{document}
\title{NeuraLeaf: Neural Parametric Leaf Models \\with Shape and Deformation Disentanglement}

\author{Yang Yang$^{1}$
\quad
Dongni Mao$^{1}$
\quad
Hiroaki Santo$^{1}$
\quad
Yasuyuki Matsushita$^{1,2}$
\quad
Fumio Okura$^1$\\
$^1$The University of Osaka \qquad $^2$Microsoft Research Asia -- Tokyo\\
{\tt\small \{yang.yang,mao.dongni,santo.hiroaki,yasumat,okura\}@ist.osaka-u.ac.jp}
}

\twocolumn[{
\maketitle
\begin{center}
    \vspace{-0.7cm}
    \fontsize{9pt}{11pt}\selectfont
    \def\svgwidth{\linewidth}
    \includegraphics[width=0.6\linewidth]{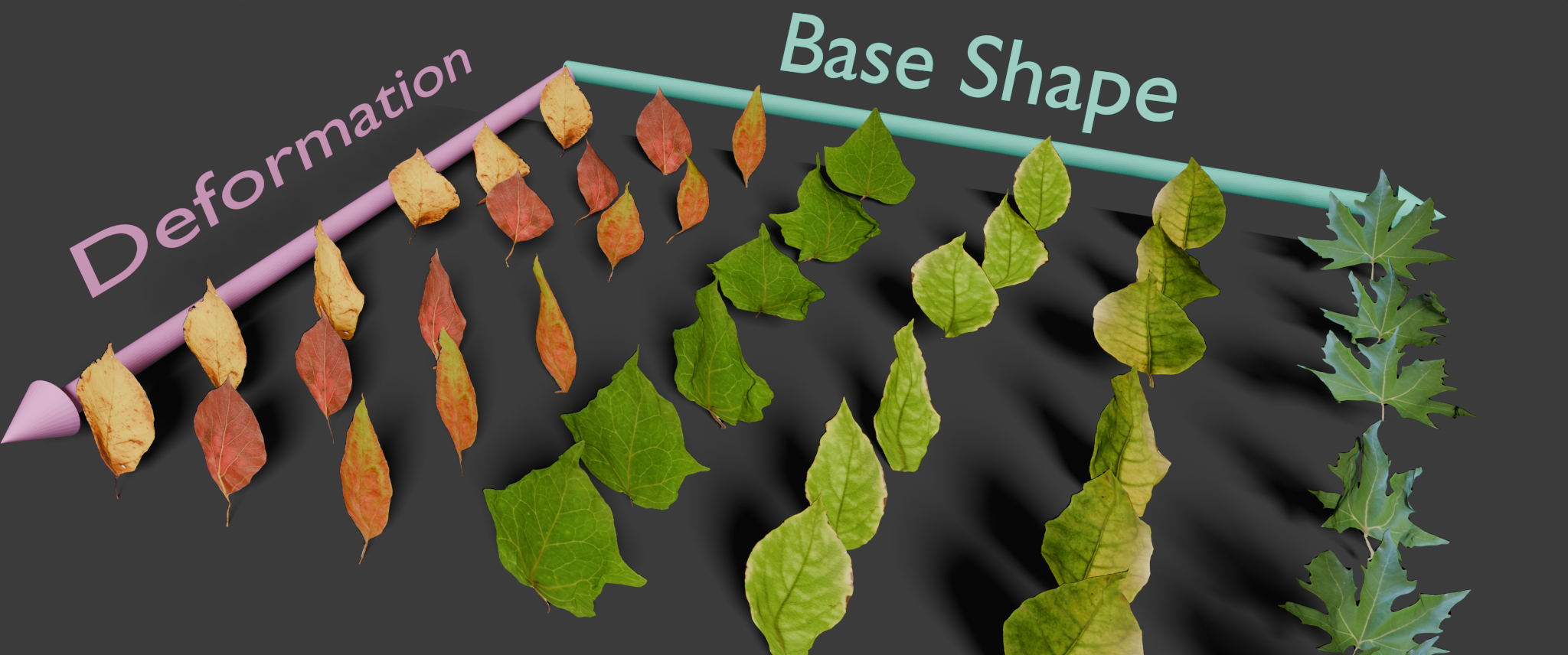}  
    \includegraphics[width=0.39\linewidth]{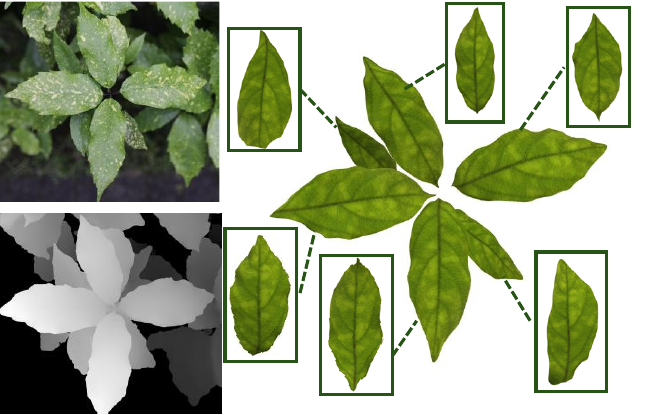} 
    \vspace{-2mm}
    \captionof{figure}{Our neural parametric model for leaves, \modelname, represents \textbf{shapes of various leaf species} and \textbf{natural 3D deformation}. Our model represents the leaves' flattened shape and their 3D deformation in disentangled latent spaces (left). Our method enables the instance-wise reconstruction of leaves via fitting to real-world observations, besides pure CG modeling (right).}  
    \label{fig:teaser}
\end{center}
}]

\begin{abstract}
We develop a neural parametric model for 3D leaves for plant modeling and reconstruction that are essential for agriculture and computer graphics. 
While neural parametric models are actively studied for humans and animals, plant leaves present unique challenges due to their diverse shapes and flexible deformation. 
To this problem, we introduce a neural parametric model for leaves, \modelname.
Capitalizing on the fact that flattened leaf shapes can be approximated as a 2D plane, \modelname disentangles the leaves' geometry into their 2D base shapes and 3D deformations.
This representation allows learning from rich sources of 2D leaf image datasets for the base shapes, and also has the advantage of simultaneously learning textures aligned with the geometry. 
To model the 3D deformation, we propose a novel skeleton-free skinning model and create a newly captured 3D leaf dataset called \datasetname. 
We show that \modelname successfully generates a wide range of leaf shapes with deformation, resulting in accurate model fitting to 3D observations like depth maps and point clouds.
Our implementation and dataset are available at \url{https://neuraleaf-yang.github.io/}.
\end{abstract}

\vspace{-3mm}
\section{Introduction}\vspace{-1mm}

\label{sec:intro}
3D leaf reconstruction is important as it provides insights into plant growth in agricultural applications and is essential for creating virtual computer graphics~(CG) assets.
For domain-specific 3D shape reconstruction, parametric 3D models have shown remarkable success, such as for human bodies (\eg, SMPL~\cite{smpl}), animals (\eg, SMAL~\cite{smal}), and even plant leaves~\cite{2013leaf, Editableleaf}. 
Although for modeling static leaves, some methods take bio-inspired approaches~\cite{wen2018leaf, runions2005modeling} using venation patterns or the nature of flattened shapes. 

Neural parametric models~(NPMs) learn the mapping between latent codes and target shapes~\cite{palafox2021npms, giebenhain2023nphm, wang2021neus, mildenhall2021nerf, muller2022instant}, showing their ability to represent detailed geometry for humans~\cite{palafox2021npms,giebenhain2023nphm, palafox2021spams,peng2021animatable, chen2021snarf} trained from large-scale human canonical shapes and their deformations.
However, the requirement of large datasets for training NPMs poses challenges for leaf modeling due to the lack of a 3D dataset containing leaves' deformations. 

To address them, we integrate leaves' biological insights with a data-driven framework. 
We propose a leaf-specific NPM representation, \modelname, leveraging the nature of leaves, \ie, the flattened leaves can be approximated as a plane.
As illustrated in \fref{fig:teaser}, we disentangle the leaves' 3D geometry into their 2D base (\ie, flattened) shape and 3D deformation represented by different latent codes. 

2D base shapes can be learned from 2D-scanned image datasets of leaves, often collected in biological studies (\eg, \cite{leafdata}, \cite{kaluzny2024laesi}), mitigating the lack of a large-scale 3D leaf dataset. 
The 2D representation offers a natural and detailed shape representation while simplifying texture modeling by treating the leaves' textures as ordinary UV maps, casting the texture generation task to an image-to-image translation task from mask image to leaf UV texture.

For the 3D deformation modeling, we newly construct a dataset, \datasetname, by acquiring real-world 3D leaf shapes in natural deformation. This dataset consists of 3D-scanned leaf shapes along with their corresponding 2D shapes in a flattened state. This is the first 3D leaf dataset focused on leaf deformation modeling, including approximately $300$ 2D-3D leaf pairs, including complex deformations and detailed geometric features like venation. Since there still remains a gap between the scale of the 2D leaf image datasets and our \datasetname due to the practical limitations in acquiring diverse 3D scans of leaves, we further introduce novel regularization to ensure shape consistency. 

Beyond data limitations, leaf deformation also presents challenges due to its high flexibility, and shape variance across species further complicates defining a common skeleton structure, making conventional skinned vertex-based models~\cite{palafox2021spams,deepsfm,tulsiani2020implicit,2000shape,wu2023magicpony,wu2022casa,yao2023hi,mihajlovic2021leap} unsuitable. We, therefore, newly introduce a skeleton-free skinning model that learns a parametric space for skinning parameters. 
Specifically, we define a set of control points on a base plane to deform the leaf surface via transformation matrices based on skinning weights, associating any leaves to a unified deformation space. 
The shape latent controls both the base shape and skinning weights, while the deformation latent predicts rigid transformations for each control point. 
We use far more control points than usual NPMs for articulated objects like humans; this is a new direction of parametric modeling of fine-grained and highly flexible deformations that are essential for accurate leaf modeling and better generalization across different leaf shapes.

We showcase applications of our \modelname for reconstruction and modeling purposes, where our method enables the \emph{leaf-wise} reconstruction from the observation as shown in \fref{fig:teaser}.
The quantitative and qualitative experiments demonstrate \modelname outperforms existing traditional parametric modeling and NPMs.

\vspace{-5mm}
\paragraph{Contributions}
The chief contributions of this paper are: 
\begin{enumerate}
\item We propose a leaf-specific NPM, \modelname, using the disentangled shape and deformation representation, which exploits the characteristics of flattened leaves. The 2D base shape representation achieves effective training from existing large 2D image datasets.
\item We newly introduce a skeleton-free skinning model for leaves, designed with many control points to capture geometric details like venation. We disentangle skinning weights and transformation matrices into two latent spaces, forming a compact representation.
\item We present a novel 3D leaf dataset \datasetname, which includes collections of base shapes and deformed leaves with their dense correspondences.
\end{enumerate}


\section{Related Work}\vspace{-1mm}

\subsection{Parametric Models}
\vspace{-1mm}

\paragraph{Traditional parametric models}
Blanz~\etal~\cite{blanz} propose an early attempt at a parametric model representing facial variations using a low-dimensional shape space extracted by principal component analysis (PCA). Along with the improvements in the 3D registration technique~\cite{arap} and large-scale data capture~\cite{anguelov2005scape,faceverse}, parametric 3D models have emerged as a leading method to model domain-specific deformable 3D shapes. 
The parametric models are studied in wide fields like human bodies~\cite{anguelov2005scape,smpl}, hands~\cite{MANO:SIGGRAPHASIA:2017}, animals~\cite{smal}, faces~\cite{FLAME}, and leaves~\cite{2013leaf,Editableleaf}.
SMPL~\cite{smpl}, a representative example of parametric human body models, combines a low-dimensional shape space with an articulated blend-skinned model and is used in various applications~\cite{Dyna:SIGGRAPH:2015,clothing1,clothing2}. 
However, such skinned vertex-based models have limited resolution and have difficulty representing non-rigid surface deformations often seen in leaves.

\begin{figure*}[t]
  \centering
   \includegraphics[width=\linewidth]{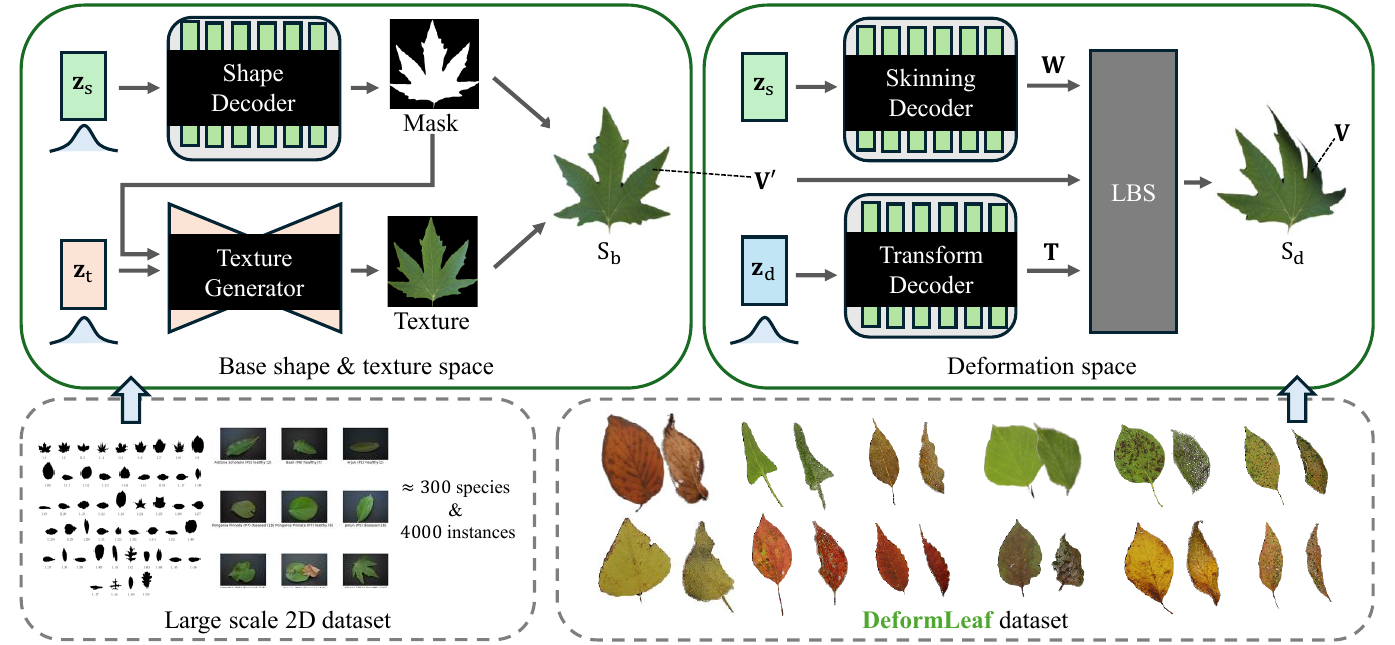}\vspace{-2mm}
  \caption{\textbf{Overview of \modelname.} We first learn the 2D base shape of leaves from a large-scale 2D dataset~\cite{plantdata}. The base shape and texture spaces are modeled by a shape decoder $f_{\theta_s}$ and texture generator $f_{\theta_t}$ conditioned on shape \& texture latent codes $\{\V{z}_{s},\V{z}_{t}\}$. Based on the learned shape space, we then train two deformation decoders to predict skinning parameters, called skinning weight decoder $f_{\theta_{w}}$ and transformation decoder $f_{\theta_d}$, which infer skinning weight of each point and transformation of each control point. We use linear blend skinning (LBS)~\cite{lbs} to deform the points.  
  We learn the deformation space from our \textbf{\datasetname} dataset containing natural deformation.}
  \vspace{-5mm}
  \label{fig:overview}
\end{figure*}

\vspace{-5mm}
\paragraph{Neural parametric models (NPMs)} 
Recently, neural implicit representations have been actively studied. DeepSDF~\cite{deepsdf}, for instance, employs a feed-forward network to predict SDF values based on query locations, conditional on a latent code. Building upon this, recent parametric models called NPMs represent the dynamic nature of objects as well as their shape using neural networks~\cite{mihajlovic2021leap,occupancyflow}.
NPMs are mainly used for modeling human shape and deformation, in which we can access large-scale 3D datasets, such as for bodies~\cite{palafox2021npms,palafox2021spams}, faces~\cite{faceverse}, and heads~\cite{giebenhain2023nphm}.

\subsection{Leaf Modeling}
\vspace{-1mm}

\paragraph{Parametric leaf modeling}
Traditionally, B-spline curves were used for representing silhouettes or veins of 2D leaves~\cite{bspline1, bspline2}, while they require tailor-made parameter sets for each group of leaves.
For leaf modeling applicable to various leaves, Barley~\etal~\cite{2013leaf} propose a parametric leaf model using a traditional PCA-based model~\cite{blanz} for leaf variation within a given flora species. 
Gaurav~\etal~\cite{Editableleaf} extend this direction for the representation of non-planar leaves by using the Bezier curve, emphasizing identifying three pivotal features of leaf geometry: midrib, left/right silhouette, and the cross-section. 
However, PCA-based models are constrained by limited shape variability, lack deformation modeling, and require large 3D datasets. 
While offering parametric control of deformation, Bezier curve-based models still cannot capture the complex and natural deformations of real leaves.
Compared to existing works, our work is the first to propose NPMs for leaves, representing natural and diverse shapes and deformations.

\vspace{-5mm}
\paragraph{Leaf deformation modeling}
Existing methods for leaf deformation modeling derive from physically-based simulations. For instance, venation skeletons and surface membranes are used to model aging leaves~\cite{hong2005shape}, allowing for interactive user-driven deformation. Similarly,~\cite{Lu2009VenationSM} animates leaf wilting due to gravity using a venation skeleton for deformation. Jeong~\etal~\cite{deformation-jeong} generate intricate wrinkles and folds in leaves by simulating inhomogeneous shrinkage, which is influenced by physical properties like loss of water.
Beyond those limited variations of deformations, our approach learns an implicit representation of deformation directly from real scanned leaf data, enabling high-fidelity and general-purpose modeling and reconstruction. 

\vspace{-5mm}
\paragraph{Bio-inspired leaf modeling} 
Although static, a bio-inspired approach~\cite{wen2018leaf} reconstructs leaf shape using a hierarchical manner using the venation and margin information. 
Similarly, \cite{runions2005modeling} generates leaves' venation patterns using leaf-specific rule sets. Our method aims to leverage the leaves' characteristics with data-driven NPMs.


\section{NeuraLeaf: Neural Parametric Leaf Model}
\label{sec:model}

\vspace{-1mm}
The overview of our disentangled representations and training strategy is illustrated in \fref{fig:overview}.

\subsection{Shape Space} \label{base}
\vspace{-1mm}
\modelname represents leaves' base shape on a plane, enabling the training of the base shape space using large-scale 2D leaf image datasets. 
We use a hybrid representation combining a 2D signed distance function (SDF) for precise boundary modeling and meshes for better capability for downstream tasks.

\vspace{-4mm}
\paragraph{Representation}
We first represent the base leaf shape in a flattened state using a neural SDF conditioned by a latent code. 
Given a latent code for a base shape (hereafter, referred to as shape latent code) $\V{z}_s\in \mathbb{R}^{N_s}$ as a condition, a multi-layer perceptron (MLP) for shape space (which we call the shape decoder $f_{\theta_s}$ maps a 2D base point $\V{x}\in \mathbb{R}^2$ to its signed distance $\Tilde{d}$ from the surface of the base shape $\mathcal{S}^*_{b}$, represented as the zero-level set of the shape MLP $f_{\theta_s}$ as
\begin{equation}
    \mathcal{S}^*_{b} =\left\{\V{x}\in \mathbb{R}^2~|~f_{\theta_s}(\V{x}, \V{z}_s)=0\right\}.
\end{equation}
We then convert the SDF into a mask $\V{I}_m$ in a differentiable manner using the sigmoid function as
\begin{equation}
    \V{I}_m(\V{x}) = \frac{1}{1+e^{-kf_{\theta_s}(\V{x})}},
\end{equation}
where $k$ is learnable to control the slope. We then apply a threshold of $0.5$ to obtain binary outputs and convert the pixels with the value $1$ into vertices on the $xy$ (\ie, $z=0$) plane. Finally, these vertices are triangulated to construct the base mesh $\mathcal{S}_b=(\mathcal{V}',\mathcal{F}')$, where $\mathcal{V}'=\{\V{v}'_i\in\real^3\}$ denotes the vertices and $\mathcal{F}'$ denotes a set of face indices.

\vspace{-4mm}
\paragraph{Training}
Thanks to the base shape spanning a plane, we can train the shape MLP $f_{\theta_s}$ using existing 2D leaf scan datasets such as~\cite{plantdata} shown in \fref{fig:overview}. 
Since the 2D scans are acquired at a controlled setup, these are straightforwardly converted to 2D silhouettes by off-the-shelf segmentation models (\eg, \cite{sam}), providing the ground-truth shape for training.
Once the ground-truth shape is obtained, we follow the common training strategy of neural implicit representations. Specifically, we sample points from a unit 2D grid along with SDF values to compute the loss functions.
SDF loss $\mathcal{L}_{\text{sdf}}$ evaluates $\ell_1$ loss between the predicted SDF $\Tilde{d}$ and the ground-truth SDF $d$ as
\begin{equation}
    \mathcal{L}_{\text{sdf}} = \|\tau(\Tilde{d}, \delta_\tau) - \tau(d, \delta_\tau)\|_1,
\end{equation}
where $\tau(d, \delta):=\min(\delta_\tau, \max(-\delta_\tau, d))$ is a truncation function controlled by $\delta_\tau$ to disregard the large SDF values. We compute the $\ell_1$ loss for both predicted SDF and resulting mask. This helps in effectively learning the differentiable conversion from SDF to mask while ensuring that the mask closely aligns with the ground-truth silhouette.
We also introduce the Eikonal regularizer~\cite{eikonal} $\mathcal{L}_{\text{eik}}$ to ensure the gradient norm of the SDF approximates. 
Furthermore, similar to~\cite{palafox2021npms}, we apply a latent regularization term $\mathcal{L}_{\text{lat}}$, with which the shape latent codes are constrained to follow a zero-mean multivariate Gaussian distribution, using a spherical covariance defined by $\sigma_s$ as
\begin{equation}
    \mathcal{L}_{\text{lat}} = \frac{1}{\sigma_s^2}\|\V{z}_s\|_2^2 .
    \label{eq:reg}
\end{equation}
We minimize the loss over all $S$ training samples to yield the shape latent codes $\{\V{z}_s^i\}_{i=1}^S$, $k$, and MLP weights $\theta_s$ as
\begin{equation}
    \underset{\substack{\theta_s,k,\{\V{z}_{s}^i\}_{i=1}^S}}{\mathrm{argmin}}\sum_{\substack{i=1 }}^S(\mathcal{L}_{\text{sdf}}+ \mathcal{L}_{\text{sil}} +\mathcal{L}_{\text{eik}} + \mathcal{L}_{\text{lat}}).
\end{equation}

\subsection{Texture Space} \label{texture}
\vspace{-1mm}

Our texture space aims to represent the leaf's appearance, aligning closely with its geometry, ensuring that the generated textures accurately correspond to the leaf's base shape.

\vspace{-4mm}
\paragraph{Representation}
We treat texture modeling as an image-to-image translation task so that an RGB image represents a texture map $\mathbf{I}_t$. Specifically, we use the CycleGAN~\cite{cyclegan} $f_{\theta_t}$ to map base mask $\mathbf{I}_m$ to its corresponding texture map $\mathbf{I}_t$ conditioned by a texture latent $\V{z}_t\in \real^{N_t}$ as 
\begin{equation}
    \V{I}_t= f_{\theta_t}(\mathbf{I}_m, \V{z}_t).
\end{equation}
Each pixel of $\mathbf{I}_t$ corresponds directly to a vertex of the base mesh, allowing straightforward UV mapping to apply the texture onto the mesh.

\vspace{-4mm}
\paragraph{Training}
We use the same datasets used in Sec~\ref{base} for training. Similar to \cite{cyclegan}, we use both GAN loss $\mathcal{L}_{\text{GAN}}$ and cycle consistency loss $\mathcal{L}_{\text{cyc}}$. 

\subsection{Deformation Space} \label{deformation}
\vspace{-1mm}

Deformation space represents the 3D deformation on the plane base mesh, trained using our \datasetname dataset containing base and deformed leaf shape pairs. We aim to learn a disentangled deformation parameter space where, given a deformation latent code, all samples, regardless of their shape, should exhibit a similar deformation pattern. 

\paragraph{Representation}
Our neural blend skinning model depends on three key components: control points, skinning weights, and rigid transformation of each control point using linear blend skinning (LBS)~\cite{lbs}. We define $K$ control points, noted as $\mathcal{C}=\{\V{c}_k\in\real^3 \mid k=1,...,K\}$, to control the deformation of each base mesh $\mathcal{S}_b$ with $N$ vertices in $\mathcal{V}'$, where $N$ varies with different base shapes. 
To represent the complex and fine-grained deformations occurring on the diverse base shapes of leaves, we use a large number of control points compared to usual articulated objects like humans. Specifically, we set the maximum number of control points $K$ to $1,000$, which is distributed uniformly across the entire UV space. 

The corresponding skinning weights are defined as $\mathcal{W} = \{w_{i,k} \mid i=1,...,N, k=1,...,K\}$, satisfying the condition where $0\leq w_{i,k}\leq{1}$ and $\sum^K_{k=1}w_{i,k}=1$. 
\begin{equation}
       \tilde{\V{v}}_i = \sum_{k=1}^K w_{i,k} \V{T}_k (\tilde{\V{v}}_i' - \tilde{\V{c}}_k), \quad \forall \V{v}_i' \in \mathcal{V}',
\end{equation}
where tildes denotes their homogeneous form, \eg, $\tilde{\V{v}}_i=[\V{v}_i,1]^\top$.
We finally get the deformed shape $\mathcal{S}_d=(\mathcal{V},\mathcal{F}')$, where $\mathcal{F}'$ are the same set of faces as in the base shape $\mathcal{S}_b$.

\vspace{1mm}
\noindent\emph{Transformation decoder: }
The purpose of the transformation decoder $f_{\theta_d}$ is to predict the rigid transformation for each control point, represented by a quaternion. Unlike previous NPMs, which predict displacement conditioned on both shape and deformation latent codes, our approach aims to learn a disentangled deformation space independent of the shape. Therefore, we condition only on the deformation latent code $\V{z}_d \in \real^{N_d}$ as 
\begin{equation}
    \mathcal{T} = f_{\theta_d}(\V{z}_d,\mathcal{C}).
\end{equation}

\vspace{1mm}
\noindent\emph{Skinning weights decoder:} 
Given a base mesh $S_b$ with its shape latent code $z_s$, the skinning weights decoder $f_{\theta_{\text{w}}}$ aims to predict the skinning weights $\mathcal{W}$ corresponding to the control points. To ensure the deformation is adapted to each specific leaf shape, the mapping is defined as 
\begin{equation}
    \mathcal{W} = f_{\theta_{w}}(\V{z}_{s},\mathcal{V}'),
\end{equation}
where $\V{z}_s$ serves as a global feature that links the skinning weights to the specific base shape. 

\vspace{-4mm}
\paragraph{Training}
The training process in the deformation space is divided into two stages to ensure that the learned deformation space generalizes well to the entire leaf base shapes.

\vspace{1mm}
\noindent\emph{First stage:} 
Unlike existing skinned shape datasets~\cite{skinningdata1, skinningdata2,skinningdata3} that provide the ground truth skinning weights and transformation matrices, our dataset is not synthetically generated and lacks the ground truth skinning parameters, \ie, we consider a correspondence-free case between the base and deformed shape. Therefore, we directly optimize the chamfer distance between predicted deformed points $\mathcal{V}$ and real points $\bar{\mathcal{V}}$ without any correspondence information.
To ensure the smoothness of the leaf surface and to prevent undesired changes to the base shape, we incorporate several common regularizers: Mesh length loss $\mathcal{L}_{\text{leng}}$ penalizes large changes in the edge of $\mathcal{S}_d$ to avoid drastic changes to the mesh geometry, and Laplacian smoothing loss $\mathcal{L}_{\text{lap}}$ encourages smoothness of the surface by minimizing differences in local curvature. 
We also use a latent regularization term $\mathcal{L}_\text{lat}$ defined similarly to \eref{eq:reg}, to constrain the latent space to a zero-mean Gaussian distribution.

\begin{figure}[t]
    \centering
     \includegraphics[width=\linewidth]{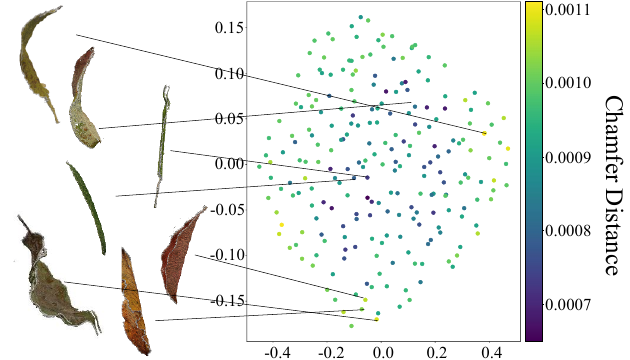}\vspace{-2mm}
    \caption{\textbf{Deformation mapping in the zero-mean latent space.} The colormap indicates the Chamfer distance between each deformed shape and its corresponding base shape. Samples closer to the center exhibit smaller deformations, while those farther from the center show greater deformations.}
    \label{fig:map}
\vspace{-3mm}
\end{figure}

Besides common constraints, we newly propose an additional regularization term, \textbf{deformation mapping loss}~$\mathcal{L}_{map}$, to align the deformation latent space. 
Inspired by the fact that the deformation space forms a zero-mean Gaussian, we measure the chamfer distance $d_{cham}$ between the base and deformed shape $(\mathcal{S}_b,\mathcal{S}_d)$. 
This loss maps the amount of deformation to the Gaussian distribution as
\begin{equation}
    \mathcal{L}_{map} = \left(\frac{{d_{cham}}(\mathcal{S}_b,\mathcal{S}_d)}{\|\V{z}_d\|_2} - \phi\right)^2,
\end{equation}
so that the latent codes move away from zero as the chamfer distance increases, as visualized in \fref{fig:map}. Here, $\phi$ is a learnable parameter that allows for an adaptable scaling factor that adjusts the penalization added by the Chamfer distance on the latent code norm.
The experiments show that the deformation space can be well trained on a relatively small dataset thanks to the disentangled latent spaces and the regularizers.

We minimize the losses over all $D$ paired samples in \datasetname to optimize the deformation latent code $\{\V{z}_d^i\}_{i=1}^D$, position of control points $\mathcal{C}$, decoder weights $\theta_d$ and $\theta_{\text{w}}$ as 
\begin{equation}
\begin{array}{ll}
    \underset{\{\V{z}_d^i\}, \mathcal{C}, \theta_{w}, \theta_d}{\mathrm{argmin}} \sum_{i=1}^D (\mathcal{L}_{\text{cham}}
    + \mathcal{L}_{\text{leng}}+\mathcal{L}_{\text{lap}} + \mathcal{L}_{\text{map}} + \mathcal{L}_{\text{lat}}).
\end{array}
\end{equation}

\vspace{1mm}
\noindent\emph{Second stage:} 
This stage aims to generalize the learned deformation space to the entire base shape space, including those without explicitly paired deformation samples. Due to the significant differences between base shapes, minimizing the Chamfer distance between dissimilar leaf pairs without altering the base shape is challenging. We thus select $m$ similar base shapes from the entire base shape space for each base shape in \datasetname based on the IoU between base masks, and we encourage similar base shapes to have similar skinning weights during training. We find that using only the Chamfer distance causes the unpaired base shape to overly conform to the deformed shape. We, therefore, introduce two additional boundary-focused constraints, \textbf{boundary length loss} $\mathcal{L}_\text{bound}$ and \textbf{face angle loss} $\mathcal{L}_{\text{ang}}$ to maintain smoothness and preserve base shapes. 

Boundary length loss aims to maintain the distance between adjacent contour vertices.
We first extract contours of the base shape $\psi$. Contour vertices on the base shape $\mathcal{S}_b$ is extracted as $\mathcal{V}'_c = \{\mathbf{v}'\in\mathcal{V}' \cap \psi\}$. Given the edge length between each pair of adjacent boundary vertices as~$\|\mathbf{v}_i' - \mathbf{v}'_{i+1}\|_2$, boundary length loss $\mathcal{L}_\text{bound}$ is defined as
\begin{equation}
\vspace{-2mm}
    \mathcal{L}_{\text{bound}}=  \sum_{\V{v}'\in\mathcal{V}'} \left( \|\mathbf{v}_i' - \mathbf{v}'_{i+1}\|_2 - \|\mathbf{v}_i - \mathbf{v}_{i+1}\|_2 \right)^2 , 
\end{equation} 
where $\mathbf{v}_i, \mathbf{v}_{i+1}\in\mathcal{V}$ represent the tentatively corresponding vertices in deformed shape $\mathcal{S}_d$ varying during training. 

Face angle loss encourages the angle around the contour vertices to become larger so that the boundaries will avoid sharp edges.
For each boundary face $f_i \in \mathcal{F}_c$ containing a contour point $\mathbf{v}_i$, the connecting edges $\mathbf{e}_1$ and $\mathbf{e}_2$ form an angle $\theta(\mathbf{e}_1, \mathbf{e}_2) $, the face angle loss $\mathcal{L}_\text{ang}$ is given by
\begin{equation}
    \mathcal{L}_{\text{ang}}= \sum_{f_i \in \mathcal{F}_c} \frac{1}{\theta(\mathbf{e}_1, \mathbf{e}_2) + \epsilon} .
\end{equation}
These two losses help to maintain the overall smoothness of the contour during the deformation process, focusing on the boundary points efficiently.

\section{DeformLeaf Dataset}
\label{sec:dataset}
\vspace{-1mm}

We construct a unique dataset of 3D deformed leaves and corresponding 2D base shapes.
In the field of human modeling, 3D/4D human datasets typically include instances of subjects performing a variety of actions, each corresponding to a canonical pose. However, in our case, while some leaves undergo rapid deformations due to dehydration, they are often uncontrollable, making it challenging to capture the deformations within a controlled environment. This fact inspires us to disentangle the shape and deformation.   

\begin{figure}[t]
	\centering
    \includegraphics[width=\linewidth]{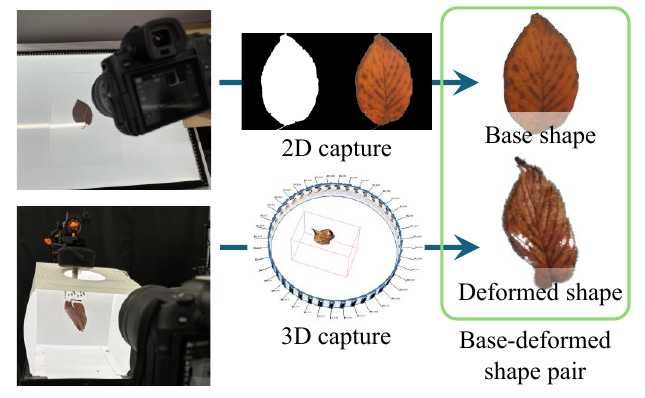}\vspace{-2mm}
    \caption{Capturing pipeline for paired base and deformed shapes.}
    \vspace{-3mm}
	\label{fig:dataset}
 \end{figure}

\vspace{-4mm}
\paragraph{Dataset description}
Our \datasetname dataset comprises approximately $300$ base-deformation leaf pairs; the capture setting is shown in \fref{fig:dataset}. 
Multi-view images of a leaf are taken from various angles to gather comprehensive visual data, and a shot from a top-down view of the base image is also taken with a light board. The 3D reconstruction process is performed by Metashape\footnote{Agisoft Metashape, \url{https://www.agisoft.com/}, last accessed on March 5, 2025.}.

\begin{figure*}[t]
    \centering
  \includegraphics[width=\linewidth]{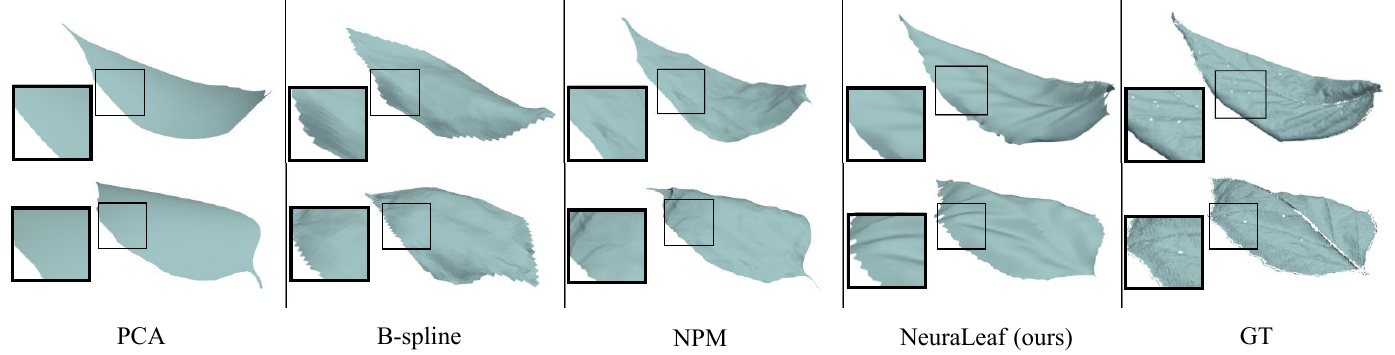}\vspace{-3mm}
  \caption{\textbf{Visual comparisons of reconstructed leaf samples.} We compare the ground truth with predicted shapes. Our method shows accurate reconstruction compared to baseline methods and has significantly more surface detail, such as leaf veins.}
  \vspace{-3mm}
    \label{fig:errormap}
\end{figure*}

\vspace{-4mm}
\paragraph{Registration} 
\label{sec:registration}
Although our \modelname does not require correspondences between the base and deformed shapes, we establish the dense correspondences in our dataset for experiments and future research.
Our pipeline involves rigid registration followed by non-rigid registration, including As-Rigid-As-Possible (ARAP)~\cite{arap} and the Coherent Point Drift (CPD)~\cite{fan2022coherent}, which gradually relaxes the rigidity in registration. See supplementary materials for details.

\section{\modelname Fitting to 3D Observations}
\label{sec:inference}
\vspace{-1mm}

Aside from pure CG modeling, \modelname can be used for reconstruction tasks by fitting our model to real-world observations.
Especially in plant science and agriculture, beyond ordinary 3D shape reconstruction, such as by multi-view stereo or RGB-D imaging, it is crucial to acquire 3D \emph{structural} information, such as topology, number, area, relationship, and direction of leaves from reconstructed shape information, in which our model helps.

The goal of NPM fitting is to find the set of latent codes, $(\V{z}_s,\V{z}_d)$ in our case, fit to the observations.  
To this end, our pipeline consists of the neural-net-based inversion of latent codes, followed by a direct optimization of the latent codes for refinement. 
We also take a similar fitting strategy for the texture latent $\V{z}_t$ but without the latent code inversion, which is detailed in the supplementary materials.

\vspace{-4mm}
\paragraph{Preliminary: Input 3D shape representation}
We represent 3D shapes in SDF grids $\V{G}\in \real^{D\times D\times D}$, providing a universal representation regardless of the input types (\eg, point clouds, mesh, and depth images). The individual leaf shapes are segmented by, such as existing segmentation methods~\cite{sam}, and roughly aligned using the rigid registration method described in Sec.~\ref{sec:registration}. 

\paragraph{Latent code inversion} \label{sec:inversion}
To obtain reasonable initialization of latent codes, we use an inversion-based approach inspired by \cite{chibane20ifnet}.
Specifically, we train encoders using 3D convolutions, $(f_{\Omega_s},f_{\Omega_d})$, which learn the mapping between the input SDF grid $\V{G}$ and the latent codes $(\V{z}_s,\V{z}_d)$ as
\begin{equation}
    \V{z}_s' = f_{\Omega_s}(\V{G}), \quad \V{z}_d' = f_{\Omega_d}(\V{G}), 
\end{equation}
where $(\V{z}_s',\V{z}_d')$ are the inverted latent codes from the observations. The encoders are simply trained with the loss function $\mathcal{L}_{inv}$ given by
\begin{equation}
    \mathcal{L}_{inv} = \|\V{z}_s'-\V{z}_s\|_2^2 +  \|\V{z}_d'-\V{z}_d\|_2^2.
 \end{equation}

\vspace{-4mm}
\paragraph{Refinement by direct optimization} \label{sec:refine}
Given the latent codes $(\V{z}_s',\V{z}_d')$ inferred by the encoders $(f_{\Omega_s},f_{\Omega_d})$, we further optimize the latent codes to fit the observation.
We compute the deformed shape $\mathcal{S}'_d$ using the estimated latent codes $(\V{z}_s',\V{z}_d')$ and the pre-trained shape, skinning weight and transformation decoders $(f_{\theta_s},f_{\theta_{w}},f_{\theta_{d}})$.
The latent codes are optimized to minimize the bi-directional chamfer loss $\mathcal{L}_{{cham}}(\mathcal{S}_d,\mathcal{S}_d')$ with the regularization term $\mathcal{L}_{reg}$ as the same manner in the training of MLPs $(f_{\theta_s}, f_{\theta_d})$ as 
\begin{align}
        \mathcal{L}_{{cham}}(\mathcal{S}_d,\mathcal{S}_d') &= \\ \nonumber
        \sum_{\V{x} \in \mathcal{S}_d} \min_{\V{x}' \in \mathcal{S}_d'} &\| \V{x}' - \V{x} \|_{2}^{2} + \sum_{\V{x}' \in \mathcal{S}'_d} \min_{\V{x} \in \mathcal{S}_d} \| \V{x}' - \V{x} \|_{2}^{2}, \\
        \mathcal{L}_{reg} &=\frac{\|\V{z}_s\|^2_2}{\sigma_b^2} + \frac{\|\V{z}_d\|^2_2}{\sigma_d^2}.
\end{align}
These loss functions are minimized through non-linear minimization using a gradient-descent method directly optimizing the latent codes $(\V{z}_s',\V{z}_d')$.



\section{Experiments}\vspace{-1mm}

We evaluate \modelname on modeling and reconstruction tasks from both quantitative and qualitative aspects. See supplementary materials for the implementation details. 

\subsection{Experiment Setup}

\paragraph{Metrics} To evaluate our reconstructions, we employed the $\ell_2$-Chamfer distance (C-$\ell_2$) and normal consistency (NC). C-$\ell_2$~[mm] is calculated between the closest point in the ground truth surface and the reconstructed surface, which evaluates point-wise discrepancies. Based on the closest correspondences, we calculate the normal consistency as the average cosine similarity between the surface normals of the ground truth and the estimated shapes.

\vspace{-4mm}
\paragraph{Baselines}
We compare our \modelname with an existing parametric leaf model using \textbf{PCA}~\cite{2013leaf}, a simplified version of the \textbf{B-spline}-based model~\cite{bspline,Editableleaf}, and a straightforward extension of an existing human-oriented \textbf{NPM}~\cite{palafox2021npms} to leaves. Supplementary materials detail the implementations for the baseline methods.

\subsection{Results}\label{ex:points}

\paragraph{Single leaf reconstruction}
Table~\ref{tab:comparison} shows a quantitative comparison on fitting to real-world single-view partial leaf point clouds, which are the test set of \datasetname containing $30$ leaves. Given the initial base shape estimated by the shape encoder $f_{\Omega_s}$, the input point cloud is rigidly aligned to the base shape at first.
According to the results, our \modelname outperforms the baseline methods, achieving higher normal consistency and also reducing chamfer distance, indicating that our proposed representation can benefit the reconstruction of the leaf surface. The visual comparisons in \fref{fig:errormap} confirm this tendency, where our method consistently achieves high-fidelity reconstruction.

\begin{table}[t]
\caption{\textbf{Quantitative comparison} between baselines on the test set of \datasetname dataset for single leaf reconstruction. The best result is highlighted in \textbf{bold}.} 
\vspace{-2mm}
 \resizebox{\linewidth}{!}{
\centering

\begin{tabular}
{l|c@{\hspace{1mm}}c@{\hspace{1mm}}c@{\hspace{1mm}}c@{\hspace{1mm}}c@{\hspace{1mm}}}
\hline
Method & $C$-$\ell_2$~[mm]  $\downarrow$ & NC $\uparrow$ & Corres-free & Temp-free &Inf.~time [s] \\ \hline
PCA~\cite{2013leaf} & $32.7$  &  $0.924$ & $\times$  &   $\times$ & 157\\
B-spline~\cite{bspline} & 26.7 &0.957 & $\surd$  & $\surd$  &  18\\
NPM~\cite{palafox2021npms} & $15.1$ &  $0.961$ & $\times$  & $\surd$ & 73\\
\modelname (ours) & $\mathbf{2.1}$ & $\mathbf{0.973}$ & $\surd$ & $\surd$ & 55\\ 
\hline
\end{tabular}
}
\vspace{-3mm}
\label{tab:comparison}
\end{table}

\vspace{-4mm}
\paragraph{Multiple leaf reconstruction}
To mimic practical use, we perform leaf reconstruction from observation of multiple leaves, which contain occlusions of leaves. We use top-view RGB-D observations of $5$ plants, where we provide segmentation masks and obtain point clouds corresponding to each leaf instance. For the ground-truth shapes, we reconstruct the plants' shapes by multi-view stereo (MVS) using many ($\geq 100$) images captured under a windless environment to accurately recover the occlusions occurring in the top-view images. The fitting process for every single leaf employs the same workflow described in Sec.~\ref{sec:inference}, while enhancing to share similar shape latent $\V{z}_s$ for the observation of the same plant species, whose details are described in the supplementary materials. For the NPM baseline, we also share the shape latent in a similar manner to our method.

\begin{figure*}[t]
    \includegraphics[width=\linewidth]{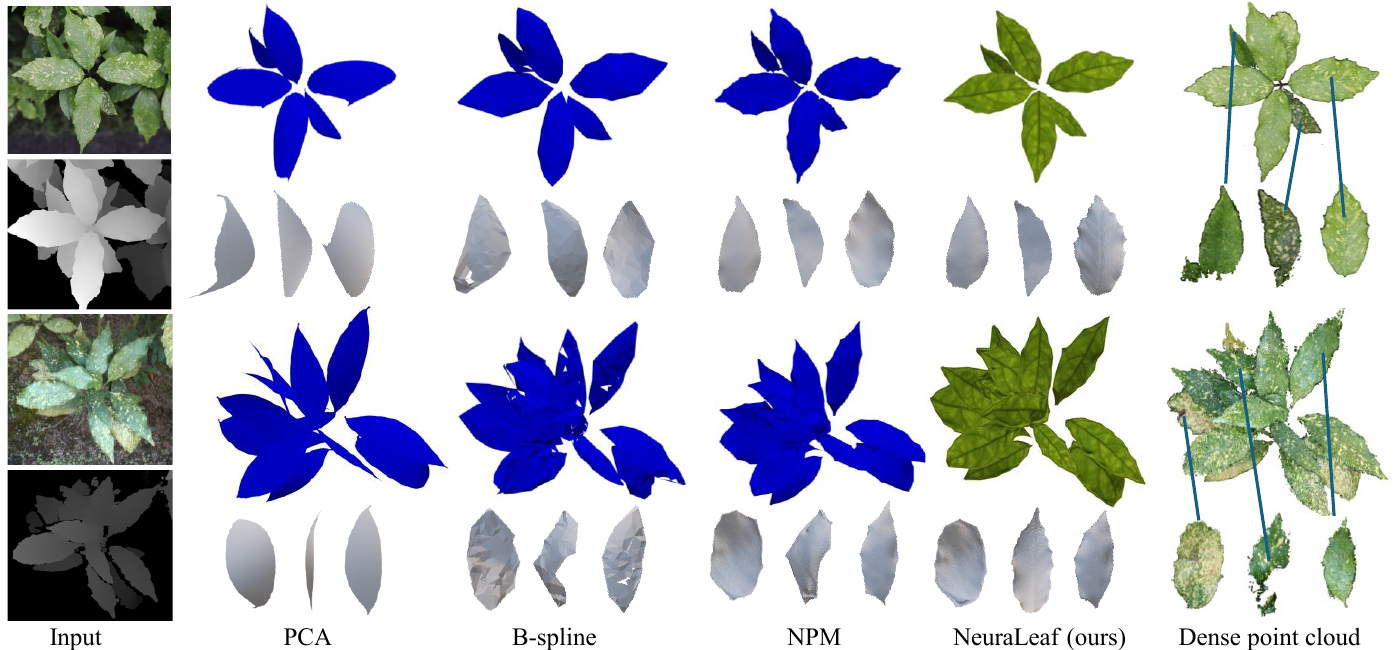}\vspace{-2mm}
    \caption{\textbf{Visual comparisons of multiple leaf reconstruction.} We use top-view RGB-D observations as input. Carefully captured multi-view images are used to create the ground truths (denoted as ``dense point cloud''). Our method achieves faithful reconstruction (\eg, vein shapes on the leaf surfaces) and better occlusion recovery (\eg, middle leaf in the bottom row).}
    \label{fig:denseleaf}
    \vspace{-4mm}
\end{figure*}

\Fref{fig:denseleaf} presents a visual example of fitting multiple leaf instances to RGB-D observation, along with the quantitative comparisons summarized in \Tref{tab:denseleaf}. 
The leaf-wise reconstruction results (see insets in \fref{fig:denseleaf}) highlight that, by sharing similar shape latent $\V{z}_s$, our method reasonably recovers leaf-wise shapes even in the portions occluded in the input RGB-D image. 
While the quantitative accuracies show an accurate reconstruction using our method, the visual examples highlight that \modelname reasonably recovers the hidden areas of leaves by sharing the shape latent among leaves.
These results are promising for practical applications in agriculture, such as measuring the leaf areas. 

\begin{table}[t]
    \centering
        \caption{\textbf{Quantitative results of multiple leaf reconstruction.}}
        \vspace{-3mm}
 \resizebox{\linewidth}{!}{
\begin{tabular}{l|@{\hspace{8mm}}c@{\hspace{8mm}}c@{\hspace{8mm}}c@{\hspace{8mm}}c@{\hspace{8mm}}}
\hline
Method & $C$-$\ell_2$~[mm]  $\downarrow$ & NC $\uparrow$ \\ \hline 
PCA~\cite{2013leaf} & 1311.3&  0.763\\
B-spline~\cite{bspline} & 159.34 & 0.652\\
NPM~\cite{palafox2021npms} (sharing shape latent) & 57.29 & 0.856 \\
\modelname (w/o sharing) & 117.14 & 0.681 \\
\modelname (full model) & $\mathbf{31.74}$ & $\mathbf{0.894}$ \\ 
\hline
\end{tabular}
}
    \label{tab:denseleaf}
    \vspace{-5mm}
\end{table}

\begin{figure}[t]
    \centering
     \includegraphics[width=\linewidth]{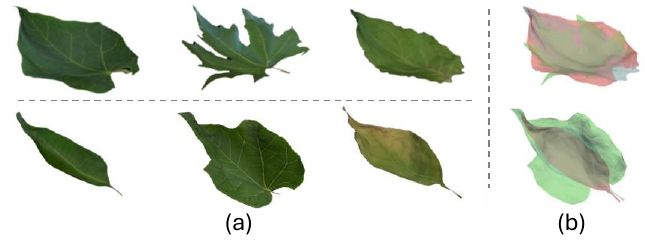}\vspace{-3mm}
    \caption{\textbf{Deformation transfer results.} Each row in (a) shows different leaf species with the same deformation latent. (b) demonstrates that each deformed leaf shape in a different color can be represented by a general surface, supporting the idea of a generalized deformation space capable of all kinds of leaf shapes. }
    \label{fig:pose transfer}
    \vspace{-2mm}
\end{figure}

\vspace{-4mm}
\paragraph{Leaf modeling applications}
Beyond reconstruction, our \modelname can be used for CG applications of leaf shape generation through generation, interpolation, and transfer of latent codes. 
An example is illustrated in \fref{fig:pose transfer}; sharing the same deformation latent code, we transform flat leaves into deformed leaves regardless of their base shapes.
This highlights the merit of our learned disentanglement deformation representation, providing deformation consistency among different leaf species. 
These applications are beneficial in CG modeling, where leaves are often modeled as simple planes and usually lack detailed deformation.
Refer to the supplementary materials and the video for other CG-related applications.

\subsection{Ablation Study}

\paragraph{Ablations of regularizer for deformation training}
We perform an ablation study to assess the effect of the deformation mapping loss $\mathcal{L}_{\text{map}}$ used for the training of the deformation space. 
Table~\ref{tab:ablation} (top) shows the results of the ablation studies and demonstrates that deformation mapping loss contributes to the improvement of shape modeling.

\begin{table}[t]
\caption{\textbf{Ablation studies.} We evaluate the chamfer distance-based regularization term $\mathcal{L}_{\text{map}}$, the latent code inversion using learned encoders, and the number of control points.} 
\vspace{-3mm}
\centering
  \resizebox{\linewidth}{!}{
\begin{tabular}{l|@{\hspace{10mm}}c@{\hspace{10mm}}c@{\hspace{10mm}}c@{\hspace{10mm}}c@{\hspace{10mm}}}
\toprule
Method & $C$-$\ell_2$~[mm]  $\downarrow$ & NC $\uparrow$ \\ 
\midrule
w/o  $\mathcal{L}_{\text{map}}$ & $5.3$  &  $0.971$  \\ 
w/o $f_{\Omega_s} $ & $9.8$ & $0.961$ \\
w/o $f_{\Omega_d}$  & $14.7$  & $0.934$ \\ 
\midrule
\midrule
\# of control points & $C$-$\ell_2$~[mm]  $\downarrow$ & NC $\uparrow$ \\
\midrule
100 points & $13.7$ & $0.932$ \\
500 points & $8.3$ & $0.958$ \\
1000 points w/o optimization & $21.4$ & $0.954$ \\
\midrule
\midrule
Ours: \modelname (1000 points) & $\mathbf{2.1}$  & $\mathbf{0.973}$   \\ 
\bottomrule
\end{tabular}
 }
\label{tab:ablation}
\vspace{-5mm}
\end{table}

\vspace{-4mm}
\paragraph{Ablations of latent code inversion} We also evaluate the ablation of latent code inversion using the encoders $f_{\Omega_s}$ and $f_{\Omega_d}$ as the initialization of latent codes for reconstruction tasks. Specifically, we initialize the shape and deformation codes using the mean of the two spaces for the cases without these encoders.
We observe that the quality of initialization by the latent code inversion significantly affects the fitting results, as shown in Table \ref{tab:ablation} (top). 

\paragraph{Ablations of control points} Since a unique characteristic of our method is using many control points, unlike NPMs for articulated objects, we conduct an ablation study on the number of control points.
\Tref{tab:ablation} (bottom) demonstrates that reducing the number of control points degrades the accuracy of reconstruction. Also, fixed (\ie, w/o optimization of) control points significantly degrades the results.


\vspace{-1mm}
\section{Conclusions}
\vspace{-1mm}
This paper has presented \modelname, a novel leaf-specific NPM that faithfully captures the unique characteristics of leaf geometry. Our method effectively disentangles a leaf's 3D structure into a 2D base shape and 3D deformation, each represented by a distinct latent code. By using large 2D-scanned image datasets and our newly acquired \datasetname dataset, \modelname learns from limited data while managing the flexibility inherent in leaf morphology. 
Through the experiments, we have shown that \modelname surpasses current traditional methods and NPMs in reconstructing leaf geometry. Also, the results show the capability of our model for real-world leaf-wise 3D reconstruction from RGB-D input.
Our contributions offer a powerful tool for biological studies, CG, and related fields. 

\vspace{-4mm}
\paragraph{Limitations}
While \modelname shows promising results in reconstructing and generating leaf shapes, there are still limitations. First, due to the limited variety of leaf species in our \datasetname dataset, the model struggles to cover certain unique types of leaf deformations. Second, although we develop a texture space, optimizing the texture code to fit the target's fine-grained appearance remains challenging.


\begin{appendix}

\renewcommand\thefigure{S\arabic{figure}}
\renewcommand\thetable{S\arabic{table}}
\renewcommand\theequation{S\arabic{equation}}

\setcounter{figure}{0}
\setcounter{table}{0}
\setcounter{equation}{0}


\maketitlesupplementary 

In this supplementary material, we first provide details for our \datasetname dataset preparation and registration methods (\sref{appendix:dataset}). We then provide more implementation details for training and evaluation, including the network architecture and implementation of the baseline methods (\sref{appendix:implementation}), as well as details for model fitting (\sref{appendix:optimization}). 

Additional experimental settings and more qualitative results are shown in \sref{appendix:results} and the supplementary video.

\section{\datasetname Dataset Preparation} \label{appendix:dataset}
In this section, we detail the dataset preparation process, corresponding to Sec.~4 of the main paper. Our dataset, \datasetname, consists of approximately $300$ base-deformation leaf pairs, where each pair contains both a 3D deformed shape and a 2D base shape with dense correspondences between them. First, we describe the extraction of the base shape from image masks in Sec.~\ref{base_extraction}. We also detail the rigid (Sec.~\ref{rigid}) and non-rigid (Sec.~\ref{non-rigid}) registration processes.
Although our method, \modelname, itself does not rely on the dense correspondences between the base and deformed shape, we consider including the correspondence information in the \datasetname dataset facilitates baseline comparisons and supports future research.

\subsection{Base Shape Extraction from Image Masks} \label{base_extraction}
Base shapes used in \modelname are extracted from 2D leaf image datasets. Given the mask region of leaves by an off-the-shelf segmentation method~\cite{sam}, we generate an SDF representation of base shapes for better compatibility with downstream tasks.

Specifically, we first calculate a 2D unsigned distance field (UDF) on the leaf's plane (\ie, $z=0$ plane). Given an image mask, the UDF is first computed using the Jump Flooding algorithm (JFA)~\cite{jumpflooding} on a 2D plane, which computes each pixel's distance to the nearest object boundary. Then, the UDFs are post-processed to ensure the correct SDF sign and normalized against the maximum distance. As we described in the main paper, base shapes can be obtained in various representations by converting the 2D SDFs to other representations, such as meshes.

\subsection{Rigid Registration} \label{rigid}
\begin{figure}[t]
    \centering
    \includegraphics[width=\linewidth]{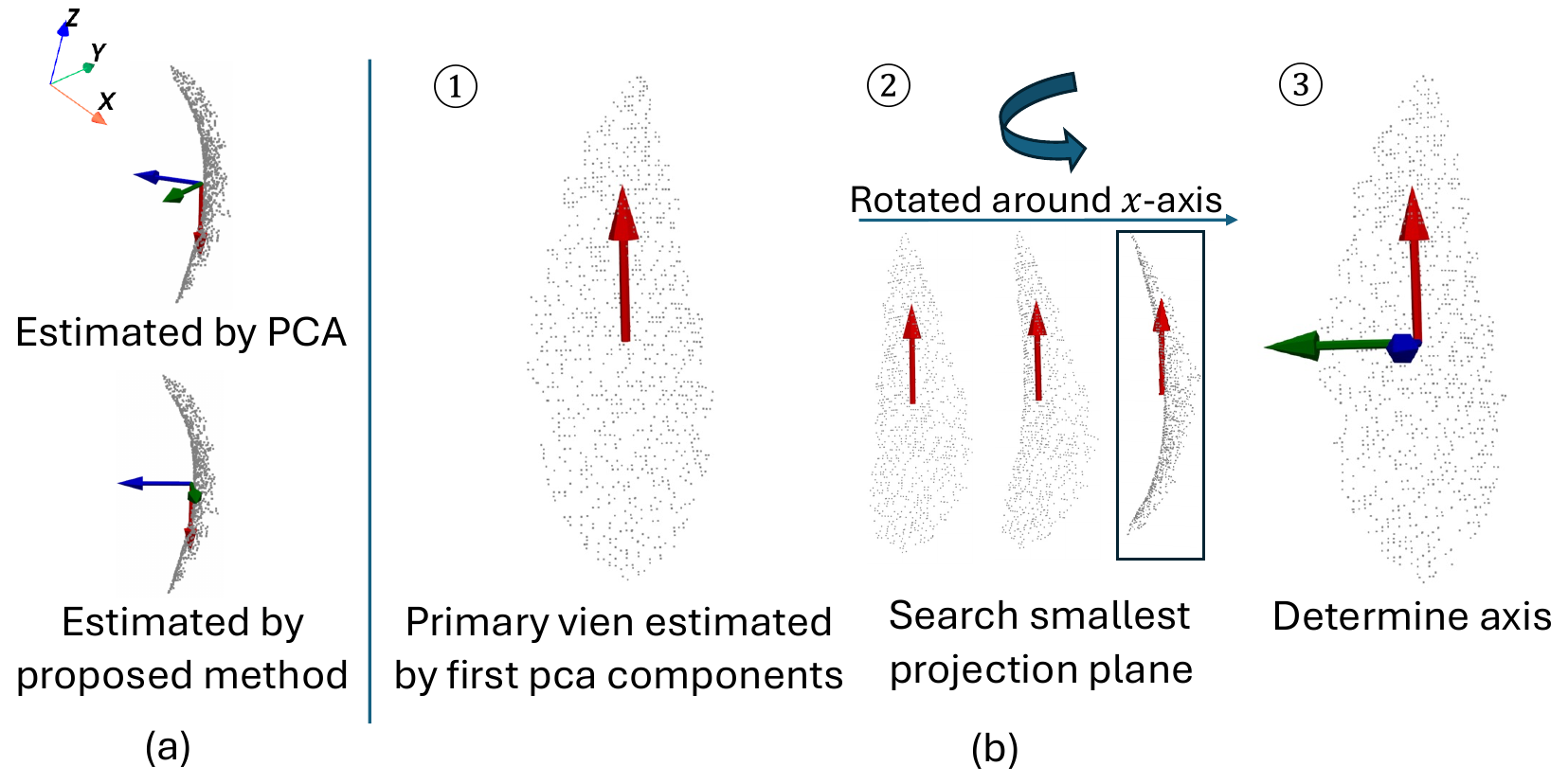}
    \caption{Rigid registration of deformed leaf shape $\mathcal{S}_d$. 
    }
    \label{fig:rigid}
\end{figure}

We first perform rigid registration for rough alignment between the base $\mathcal{S}_b$ and deformed $\mathcal{S}_d$ shapes.
We take a similar but slightly different approach from the usual PCA-based alignment. 
The key to leaf pose estimation is to confirm the primary and secondary veins of the surface. Firstly, we compute the largest principal component of the vertices of the deformed shape to determine the primary vein ($x$-axis). 
Although conventional PCA-based alignment methods~\cite{leafpca} use the second principal component as the secondary vein ($y$-axis), as leaf deformation increases, the shape variance around the secondary vein becomes large, as shown in \fref{fig:rigid}(a).
Instead, we seek the secondary vein ($y$-axis) as the direction perpendicular to the $x$-axis, where the projected area of the 3D shape is minimal, as in \fref{fig:rigid}(b).

\subsection{Non-rigid Registration} \label{non-rigid}
As mentioned in Sec.~4 in the main paper, we perform a non-rigid registration pipeline using the As-rigid-as-possible (ARAP) registration followed by the Coherent Point Drift (CPD) registration. Since the leaves have flexible deformation, yielding a reasonable registration by a single method is often challenging. Therefore, we combine two registration methods to relax the degrees of freedom in the deformation gradually: rigid alignment, ARAP assuming a locally rigid shape, and CPD as non-rigid registration.

\paragraph{As-rigid-as-possible (ARAP) registration}
In human and animal models, the deformations rely on their skeletal structures with hinged joints; \ie, the deformation can be characterized as a rigid motion connected at joints. ARAP allows the model to bend or rotate around these joints while keeping the local shapes. 
Since leaves lack the hinged joints of the skeleton, we artificially designate the joint points (hereafter called keypoints) by uniformly sampling along both the base and the deformed leaves' contour. By sampling the same number of points, we yield the correspondences between keypoints on the base shape and the scanned (\ie, deformed) shapes.
 
Given the set of keypoints, $\mathcal{V}=\{\mathbf{v}_i\in\real^3\}_i$ and $\mathcal{V}'=\{\mathbf{v}'_i\in\real^3\}_i$, defined on the base shape and deformed (\ie, scanned) shape respectively, ARAP optimizes the vertex-specific offsets $\left\{\mathbf{\delta}_i\in\real^3\right\}$ and rotation $\left\{\mathbf{R}_i\in\real^{3\times 3}\right\}$. Specifically, we minimize the following energy function:
\begin{equation}
    \sum_{i} \sum_{\mathbf{u}\in\mathcal{N}_{\mathbf{v}_i}}(d(\hat{\textbf{v}}_i - \textbf{v}'_i) + \|(\hat{\textbf{v}}_i-\hat{\textbf{u}})-\mathbf{R}_i(\textbf{v}_i-\textbf{u})\|_2^2 ),
\end{equation}
where $\hat{\textbf{v}}_i= \textbf{v}_i + \delta_i$, $d(\hat{\textbf{v}}_i - \mathbf{v}'_i)$ is the mean point-to-point distance from $\hat{\textbf{v}}_i$ to its nearest neighbors in $\mathcal{V}'$, and $\mathcal{N}_{\mathbf{v}_i}$ denotes the nearest neighboring vertices to ${\mathbf{v}_i}$.

\paragraph{Coherent point drift (CPD) registration}
We use a state-of-the-art implementation of CPD~\cite{fan2022coherent}, which extends the original Euclidean CPD to non-Euclidean domains using the Laplace-Beltrami operator.
This method addresses the challenge of a large search space using the nearest-neighbor descriptor matching to find confident correspondences. The probabilistic model of CPD is refined to manage outlier-contaminated initial correspondence sets effectively, employing an Expectation-Maximization (EM) approach for improved robustness and efficiency. See more details in \cite{fan2022coherent}.

\section{Implementation Details} \label{appendix:implementation}

We train the \modelname model on a single RTX A6000 GPU using our \datasetname dataset. 
The dimensions of shape latent $N_s$, texture latent $N_t$, and deformation latent $N_d$ are set to $256$. For shape decoder $f_{\theta_s}$, we apply the same architecture introduced in the Implicit Differentiable Renderer (IDR)~\cite{IDR} and train our models for $3,000$ epochs. For the skinning decoder and transformation decoder, we use the network introduced in \cite{skinningdata2}.
We use a learning rate of $10^{-3}$ for the networks and the latent codes. We implement a learning rate decay every $1,000$ epochs. 
Similar to \cite{palafox2021npms}, latent inversion encoders $(f_{\Omega_s},f_{\Omega_d})$ receive back-projected observation in the form of a partial SDF grid and output the latent code estimation through a series of 3D convolution layers and a final fully-connected layer. We hereafter detail the implementations further.

\subsection{Architecture of Decoders}
In this section, we provide an overview of the different networks used in our framework, each serving a specific role in capturing leaf shape, deformation, and texture. Specifically, our framework includes the following decoders: a \textbf{Shape decoder} $f_{\theta_s}$ for base shape representation, a \textbf{Texture generator} $f_{\theta_t}$ for modeling the leaf texture, a \textbf{Skinning Weights decoder} $f_{\theta_w}$ for predicting the influence weights of control points, and a \textbf{Transformation decoder} $f_{\theta_d}$ for calculating the transformations at each control point to achieve the desired deformation.
\paragraph{Shape decoder}
Shape decoder $f_{\theta_s}$ is an MLP that consists of $8$ layers, with a dimension of hidden layers of $512$ and a single skip connection from the input to the middle layer, which is similar to an existing NPM~\cite{chibane20ifnet}. The initialization strategy of parameters of $f_{\theta_s}$ is the same as \cite{Atzmon_2020_CVPR}, so that $f_{\theta_s}$ produces an approximate SDF of a unit sphere. 

\paragraph{Texture generator}
Similar to ~\cite{cyclegan}, the architecture of texture generator $f_{\theta_t}$ contains three convolution layers, multiple residual blocks, and two fractionally-strided convolution layers with stride $\frac{1}{2}$. We apply $9$ blocks to $256 \times 256$ images and also use instance normalization. For the discriminator network, we apply a $70 \times 70$ PatchGAN to discriminate whether the overlapping image patches are real or fake.

\paragraph{Skinning weights decoder}
Skinning weights decoder $f_{\theta_{w}}$ uses the architecture introduced in ~\cite{saito2021scanimate}, which is an MLP with intermediate layer shape $(256,256,256,1000)$ with a skip connection from the input feature to the 2nd layer, and non-linear activations using LeakyReLU except the last layer uses softmax layer to obtain normalized outputs. As input, we take the Cartesian coordinates of a base point sampled from base shape $\mathcal{S}_b$, which is encoded into a high-dimensional feature using positional encoding ~\cite{rahaman2019spectral}, incorporating up to 8th-order Fourier features. 

\paragraph{Transformation decoder}
The transformation decoder $f_{\theta_d}$ follows a similar network architecture as the skinning weights decoder with an intermediate layer shape (256,256,256,7), where the first four dimensions represent a quaternion for rotation and the remaining three dimensions represent translation. The input to the network consists of points sampled from the control points $\mathcal{C}$ and also followed by positional encoding.

\subsection{Training Details}
We use Adam optimizer for $f_{\theta_s}$, $f_{\theta_{w}}$ and $f_{\theta_d}$ and also apply a learning rate decay factor of $0.5$ every $500$ epochs. For training of base shape space, the SDF truncation is set to $\delta_\tau=0.01$ and spherical covariance defined by $\sigma_s$ is $10$. For deformation latent codes, the spherical covariance defined by $\sigma_d$ is also $10$. Both shape latent codes $\{\mathbf{z}_s^i\}_{i=1}^S$ and deformation latent codes $\{\mathbf{z}_d^j\}_{j=1}^D$ are initialized randomly from $\mathcal{N}(0,0.001)$.

\subsection{Implementation of Baseline Methods}
\paragraph{PCA-based parametric model~\cite{2013leaf}} We evaluate the parametric leaf model proposed by Barley~\etal~\cite{2013leaf}. 
This PCA-based approach needs dense correspondence among all shapes within the dataset.
To accommodate this requirement, we partition our base shape dataset into several groups based on shape characteristics. Each group's shapes are registered to a template shape, then the parametric 
space is obtained based on BFM~\cite{BFM}. The template shapes are selected from \datasetname rather than using low-polygon leaf meshes, ensuring that the PCA baseline and other methods are compared at the same mesh resolution.
For deformation modeling, we apply the same non-rigid fitting pipeline introduced in ~\cite{2013leaf} and optimize the shape parameter simultaneously during inference.

\paragraph{Neural parametric model~\cite{palafox2021npms}} We compare our method with a straightforward extension of the existing NPM~\cite{palafox2021npms}, trained using our \datasetname dataset.
Deformation modeling in this baseline is based on the neural displacement field. Thus, we use the dense correspondences in \datasetname for training. The training pipeline follows the paradigm in \cite{palafox2021npms}.


\begin{figure*}[tp]
    \centering
    \includegraphics[width=\linewidth]{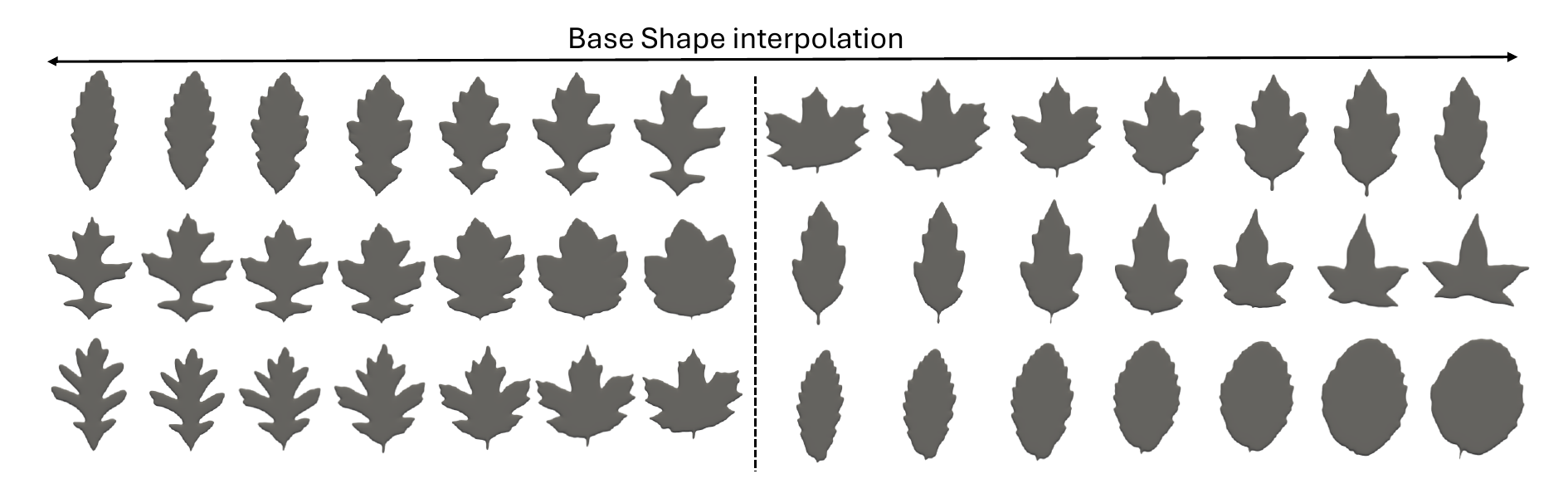}
    \caption{Interpolation between diverse base shapes, showing the smoothness of our base shape latent space. }
    \label{fig:interpolation}
\end{figure*}
\begin{figure*}[tp]
    \centering
   \includegraphics[width=\linewidth]{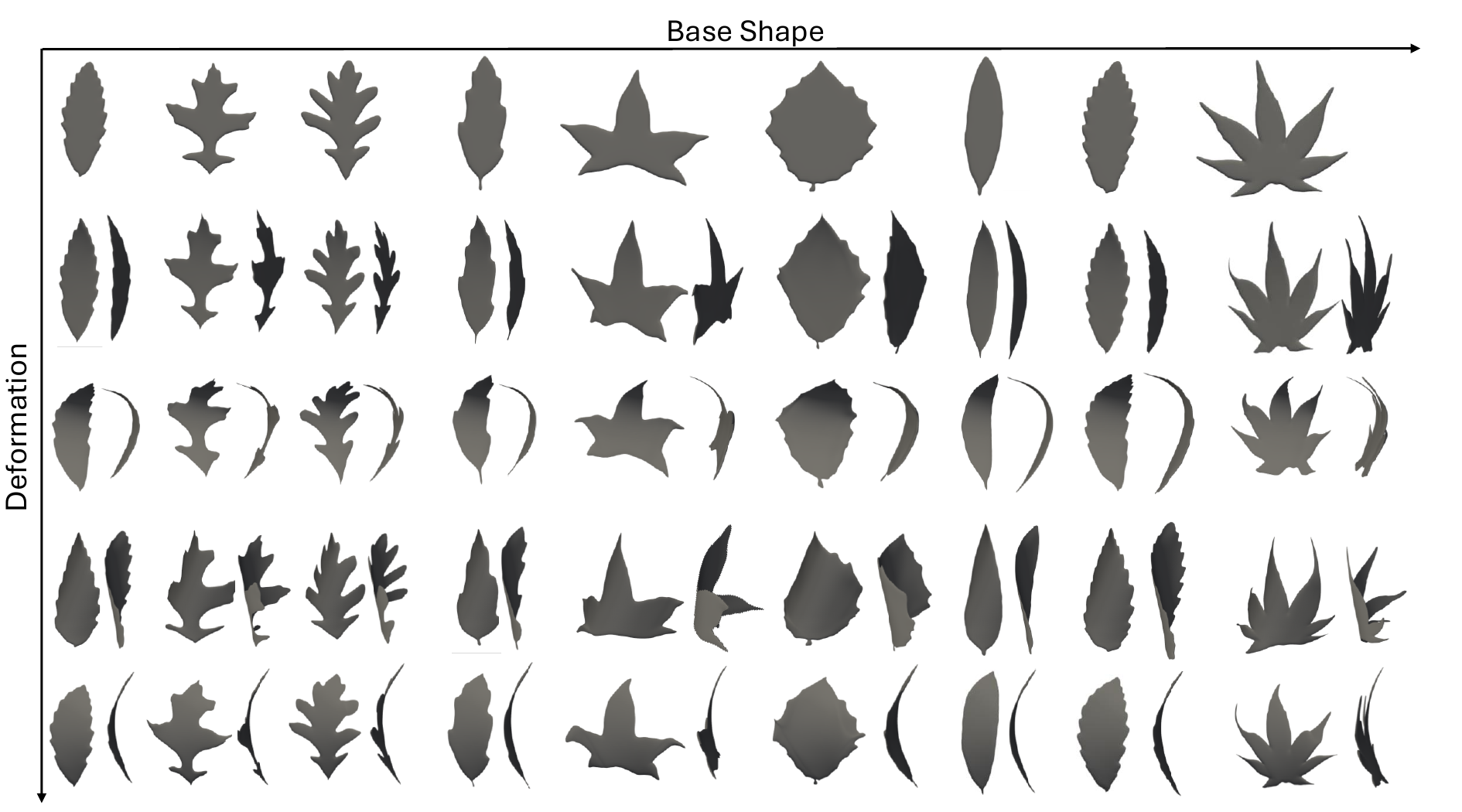}
    \caption{Deformation among different kinds of base shapes. Each row shows the results using the same deformation latent codes.}
    \label{fig:deformation}
\end{figure*}

\begin{figure}[tp]
    \centering
    \includegraphics[width=\linewidth]{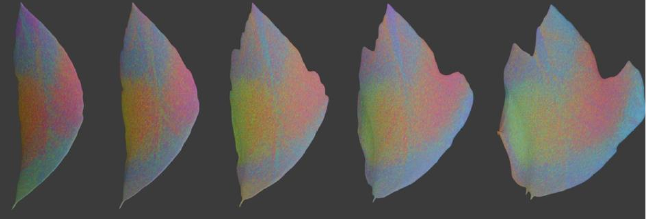}
    \caption{Visualization of skinning weights during shape interpolation.}\vspace{-5mm}
    \label{fig:skinning}
\end{figure}

\begin{figure}[tp]
  \centering
  \includegraphics[width=\linewidth]{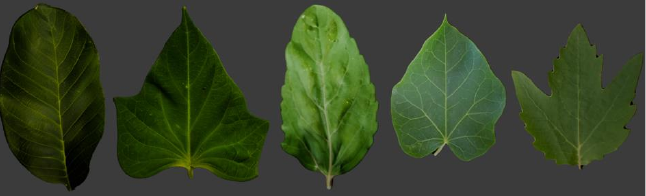}
  \caption{Shape generation via random sampling of shape latent.}\vspace{5mm}
   \label{fig:random}
\end{figure}

\section{Model Fitting to 3D Obeservations} \label{appendix:optimization}

\subsection{Details for Latent Code Inversion}
As discussed in the main paper, we use two 3D convolutional networks $f_{\Omega_s}$ and $f_{\Omega_d}$ to map the back-projected SDF grid $\mathbf{G}\in \mathbb{R}^{128\times128\times 128}$ to the initial latent code $\mathbf{z}_s$ and $\mathbf{z}_d$ at the inference stage. The convolutions are followed by a downsampling that creates growing receptive fields and channels with shrinking resolution, similar to common practices in 2D convolution. By applying this process recursively $n$ times to input grid $\mathbf{G}$, it creates multi-scale deep feature grids $\{\mathbf{G}_1, ..., \mathbf{G}_n\}$, each with a decreasing resolution $\mathbf{G}_k\in \mathbb{R}^{K\times K\times K}, \text{where } K=\frac{128}{2^k-1}$. Feature grids $\mathbf{G}_k$ at early stages capture high frequencies, indicating fine shape details, whereas feature grids at later stages capture the global structure of the data.

\subsection{Details for Direct Optimization}
After initial latent code $\textbf{z}_s$ and $\textbf{z}_d$ predicted, we optimize for $\textbf{z}_s$ and $\textbf{z}_d$ by minimizing Eqs.~(13) and (14) in the main paper. We use Adam optimizer and learning rate of $1 \times 10^{-3}$ for both $\textbf{z}_s$ and $\textbf{z}_d$. We optimize for a total of 300 iterations and perform learning rate decay by a factor of $0.5$ every $50$ iterations. 

\subsection{Texture Latent Estimation}
In the limitation section of the main paper, we mention that our method has shortcomings in fitting the texture, as the texture latent code cannot be fitted to the input through fine-tuning. Therefore, we take the mean of the pre-trained texture space as the initial value, which is then updated.

\subsection{Details for Multiple Leaf Reconstruction }
To begin the reconstruction process, each leaf instance is initially processed by the shape encoder $f_{\Omega_s}$ to estimate the shape latent code $\V{z}_s$. However, due to occlusion or overlapping regions among leaves, some of the estimated shape codes might not accurately represent the true leaf shape. To identify a representative anchor latent code in the presence of potential outliers, we first compute the similarity between latent codes using Euclidean distance and perform clustering. Specifically, we employ K-means clustering to divide the latent codes into $K=3$ clusters and select the cluster with the largest number of latent codes as the main cluster, assuming that this cluster represents the normal shape. Once the main cluster is identified, we choose an anchor latent code $\V{z}_s^{anc}$ by selecting the code closest to the cluster center. $\V{z}_s^{anc}$ serves as the reference during optimization. Formally, we introduce a regularization term $\mathcal{L}_{anc}$ defined as 
\begin{equation}
    \mathcal{L}_{anc} = \|\V{z}_s\ - \V{z}_s^{anc}\|_2,
\end{equation}
ensuring that the shape latent codes for each leaf instance do not deviate significantly from the reference shape.

\begin{figure*}[tp]
    \centering
   \includegraphics[width=\linewidth]{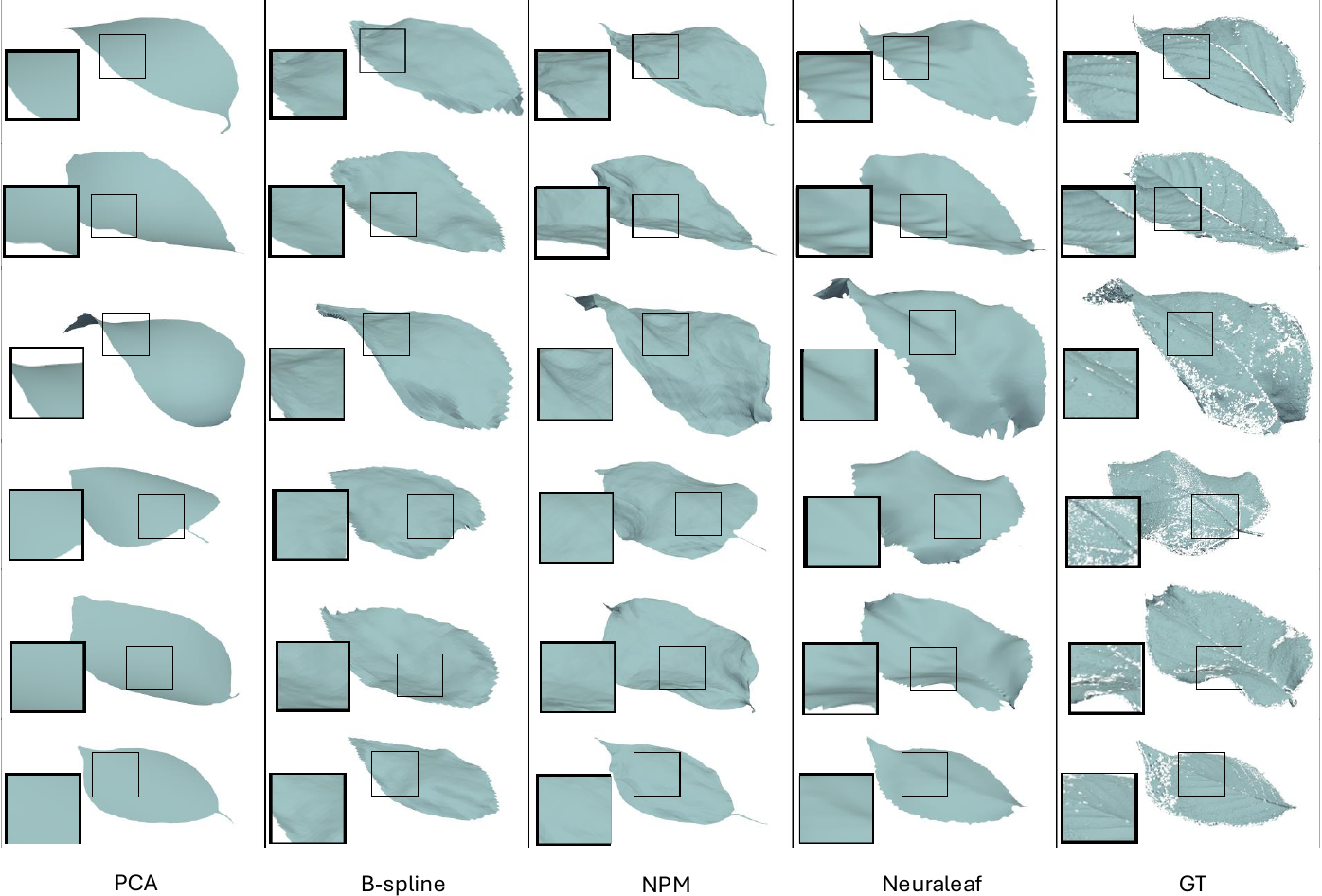}
    \caption{Additional comparison of single-leaf reconstruction.}
    \label{fig:compare}
\vspace{-3mm}
\end{figure*}

\begin{figure}[tp]
    \centering
    \includegraphics[width=\linewidth]{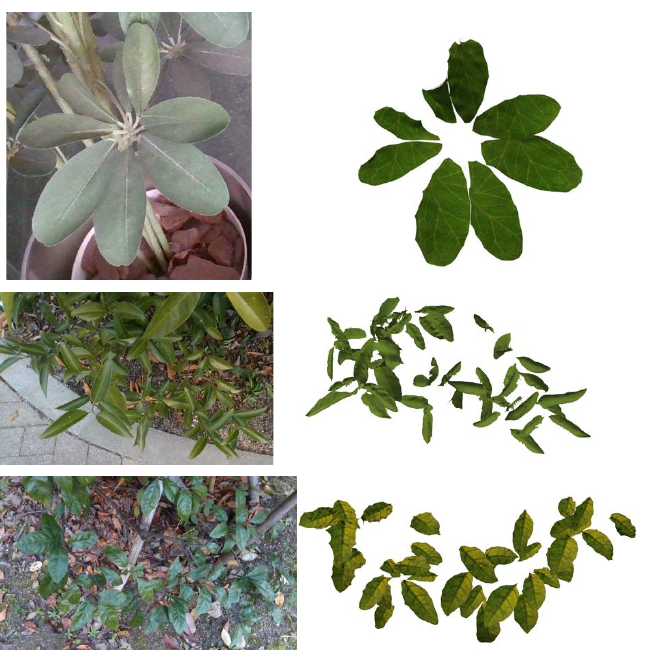}
    \caption{Additional result of multiple leaf reconstruction.}\vspace{-8mm}%
    \label{fig:supp_denseleaf}
\end{figure}

\begin{figure}[tp]
    \centering
    \includegraphics[width=\linewidth]{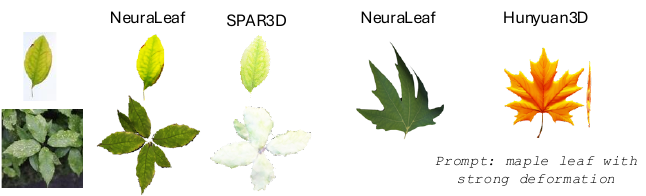}
    \caption{Comparisons with 3D model generation methods.}
    \label{fig:gene}
\vspace{-3mm}
\end{figure}

\begin{table}[tp]
\centering
\caption{Additional ablation studies with $\mathcal{L}_\text{bound}$ and $\mathcal{L}_\text{ang}$.}
   \resizebox{\linewidth}{!}{
\begin{tabular}{l|@{\hspace{10mm}}c@{\hspace{10mm}}c@{\hspace{10mm}}c@{\hspace{10mm}}c@{\hspace{10mm}}}
\hline
Method & $C$-$\ell_2$~[mm]  $\downarrow$ & NC $\uparrow$ \\ \hline
w/o $\mathcal{L}_\text{bound} $ & 9.4 & 0.971 \\
w/o ${\mathcal{L}_\text{ang}}$  &  4.3 & 0.953 \\ \hline
\modelname(full model) & $\mathbf{2.1}$  & $\mathbf{0.973}$   \\ 
\hline
\end{tabular}
 }
\label{tab:supp_ablation}
\end{table}

\section{More Results and Analysis} \label{appendix:results}
We also strongly encourage readers to refer to the supplementary video for more intuitive visualizations.

\vspace{-3mm}
\paragraph{Shape interpolation}
We show interpolation results between base shapes with distinct differences in \fref{fig:interpolation}, showing the smoothness and compactness of our base latent space, which is useful for CG leaf shape modeling using \modelname. 

\vspace{-3mm}
\paragraph{More generation examples}
We also show different samples in the same deformation pattern (\ie, deformed by the same deform code $\textbf{z}_d$) in \fref{fig:deformation}, to evaluate the deformation consistency among different base leaf shapes. 

\vspace{-3mm}
\paragraph{Visualization of skinning weights}
We also show the results of interpolating shape codes with fixed deformation in \fref{fig:skinning} along with their skinning weights.
Although skinning weights are shape-dependent, they are mostly consistent across base shapes, allowing the model to adapt to unseen leaf shapes.

\vspace{-3mm}
\paragraph{Random shape generation}
\modelname also allows the leaf shape generation by random sampling in the shape space, as shown in \fref{fig:random}, showing that the shape latent is reasonably constructed.

\vspace{-3mm}
\paragraph{More reconstruction examples}
More reconstruction results of single and multiple leaves are shown in \fref{fig:compare} and \fref{fig:supp_denseleaf}, respectively, to further emphasize \modelname's performance in the leaf reconstruction task.

\vspace{-3mm}
\paragraph{Comparisons with 3D generation methods}
Although the domain is a bit different, we further compare \modelname to SOTA image-to-3D~\cite{spar3d} and text-to-3D~\cite{hunyuan3d} models. The results in \fref{fig:gene} show that ours produces more faithful leaf geometries, deformation, and texture.

\vspace{-3mm}
\paragraph{More ablations}
We provide additional ablation studies specifically targeting boundary length loss and face angle loss in Table~\ref{tab:supp_ablation}, demonstrating that they can help avoid overfitting to local optima.

\end{appendix}

\section*{Acknowledgments}
\vspace{-1mm}
This work was supported in part by the JSPS KAKENHI JP23H05491, JP25K03140, and JST FOREST JPMJFR206F.

{
    \small
    \bibliographystyle{ieeenat_fullname}
    \bibliography{main}
}

\end{document}